\documentclass[]{spie}  %>>> use for US letter paper

\usepackage{amsmath,amsfonts,amssymb}
\usepackage{subcaption}
\usepackage{graphicx}
\usepackage{cite} 
\usepackage{epsfig}
\usepackage{nccmath}
\usepackage{xcolor,colortbl}
\usepackage{tabularx}
\usepackage{relsize}
\usepackage{pifont}
\usepackage{booktabs} 
\usepackage{multirow}
\usepackage{multicol}
\usepackage{adjustbox}
\usepackage{float}
\usepackage{makecell}
\usepackage{tabu}
\usepackage[colorlinks=true, allcolors=blue]{hyperref}
\usepackage[capitalize]{cleveref}
\usepackage{algorithm}
\usepackage{subcaption}
\title{MedFoundationHub: A Lightweight and Secure Toolkit for Deploying Medical Vision Language Foundation Models}

\author[a]{Xiao Li}
\author[a]{Yanfan Zhu}
\author[b]{Ruining Deng}
\author[c]{Wei-Qi Wei}
\author[c]{Yu Wang}
\author[c]{Shilin Zhao}
\author[d]{Yaohong Wang}
\author[c]{Haichun Yang}
\author[a]{Yuankai Huo}

\affil[a]{Vanderbilt University, Nashville TN 37235, USA}
\affil[b]{Weill Cornell Medicine, New York, NY 10021, USA}
\affil[c]{Vanderbilt University Medical Center, Nashville TN 37232, USA}
\affil[d]{UT MD Anderson Cancer Center,TX 77030, USA}

\authorinfo{Further author information: (Send correspondence to Yuankai Huo)\\
E-mail: yuankai.huo@vanderbilt.edu}

\begin{document}
\maketitle

\begin{abstract}
Recent advances in medical vision-language models (VLMs) open up remarkable opportunities for clinical applications such as automated report generation, physician copilots, and uncertainty quantification. Despite their promise, medical VLMs raise serious security concerns. These include the risk of Protected Health Information (PHI) exposure, data leakage, and vulnerability to cyberthreats, concerns that are especially critical in hospital environments. Even when adopted for research or non-clinical purposes, healthcare organizations must exercise caution and implement safeguards. To address these challenges, we present MedFoundationHub, a graphical user interface (GUI) toolkit that: (1) enables physicians to manually select and use different models without programming expertise, (2) supports engineers in efficiently deploying medical VLMs in a plug-and-play fashion, with seamless integration of Hugging Face open-source models, and (3) ensures privacy-preserving inference through Docker-orchestrated, operating system agnostic deployment. MedFoundationHub requires only an offline local workstation equipped with a single NVIDIA A6000 GPU, making it both secure and accessible within the typical resources of academic research labs. To evaluate current capabilities, we engaged board-certified pathologists to deploy and assess five state-of-the-art VLMs (Google-MedGemma3-4B, Qwen2-VL-7B-Instruct, Qwen2.5-VL-7B-Instruct, and LLaVA-1.5-7B/13B). Expert evaluation covered colon cases and renal cases, yielding 1,015 clinician–model scoring events. These assessments revealed recurring limitations, including off-target answers, vague reasoning, and inconsistent pathology terminology.
\end{abstract}

\section{Introduction}
The rapid development of large vision--language models (VLMs) has created unprecedented opportunities for advancing medical artificial intelligence (AI). By jointly leveraging visual and textual representations, medical VLMs have demonstrated promise in a wide range of applications, including automated diagnostic report generation, interactive copilots for clinicians, decision support under uncertainty, and knowledge-assisted question answering. These capabilities have the potential to improve efficiency, reduce cognitive burden, and ultimately enhance the quality of patient care. \cite{biogpt}\cite{ medvlm} \cite{healthbench} At the same time, progress in biomedical AI shows a clear trend toward domain-specific and integrative foundation models. \cite{x-rem} \cite{omiclip} \cite{brainmvp} \cite{polypath}

However, the deployment of such models in real-world healthcare environments remains fraught with challenges. Chief among these are concerns related to \textbf{data privacy and security}. Because medical VLMs may process or infer from clinical data, any mishandling of \textit{Protected Health Information (PHI)} poses serious risks of privacy violations and legal non-compliance, especially under regulations such as HIPAA in the United States \cite{hipaa_security}. Furthermore, as recent reports of cyberattacks on healthcare systems illustrate, AI systems that lack robust security safeguards may introduce new vulnerabilities, including the risk of data leakage, unauthorized access, or model inversion attacks \cite{cyberattack_healthcare}. These risks are compounded when models are hosted on third-party servers, making many cloud-based solutions unsuitable for hospital and research settings. 

Beyond privacy and security, \textbf{practical barriers} further limit adoption. Current medical VLMs often require significant engineering expertise for installation, model integration, and deployment\cite{guide, hpm, vlm}. This complexity restricts their accessibility to physicians and researchers who may benefit most from their use. Moreover, the computational demands of many large foundation models pose additional constraints, as clinical and academic institutions often lack the infrastructure to host and maintain massive GPU clusters\cite{gpu1, gpu2}. Thus, there exists a critical need for solutions that are both \textbf{secure} and \textbf{lightweight}, lowering the barrier to entry for research labs and clinical environments alike.

Modern VLM/LLM pipelines for medical AI increasingly rely on contrastive alignment between images and text to obtain reusable embeddings for retrieval, captioning, and downstream diagnosis. General frameworks such as CLIP learn image--text correspondences at scale and enable zero-shot transfer \cite{radford2021clip}; pathology-specialized variants (e.g., PLIP and CONCH) curate domain captions and reports to inject histology lexicon and improve discriminative power \cite{plip, conch}. On the generative side, slide-level pathology models like PRISM \cite{prism,prism2}.

To bridging these translation gaps, we propose \textbf{MedFoundationHub}, a lightweight and privacy-preserving toolkit for deploying medical VLMs. MedFoundationHub is designed with three key principles: (1) \textbf{physician-friendly interface}---providing a graphical user interface (GUI) that allows physicians to interact with multiple models without requiring programming skills, (2) \textbf{plug-and-play deployment}---offering engineers a plug-and-play architecture that supports automated integration of open-source VLMs from Hugging Face, and (3) \textbf{security by design}---ensuring all inference occurs on local, Docker-orchestrated workstations to protect sensitive data. Importantly, the system runs efficiently on a single NVIDIA A6000 GPU, enabling adoption in typical research lab settings without the need for specialized infrastructure. 

To assess the emerging capabilities of frontier vision–language models in pathology, we conducted structured evaluations with board-certified pathologists across five representative systems: Google-MedGemma3-4B\cite{medgemma3}, Qwen2-VL-7B-Instruct\cite{qwen2_vl}, Qwen2.5-VL-7B-Instruct\cite{qwen2_5_vl}, and LLaVA-1.5-7B/13B\cite{liu2023visual}. 
Expert-reviewed cases were drawn from the PathologyOutlines Colon QA\cite{pathologyoutlines_colonqa}, together with renal pathology resources including Arkana Labs\cite{arkanlabs_renalqa}, the AJKD Atlas of Renal Pathology\cite{ajkd_atlas_tma}, NephSIM\cite{nephsim_jgme2019}, and other curated sources yielding more than one thousand clinician–model interaction scores. 
Despite encouraging advances, our evaluations show that these models still drift off-topic, reason inconsistently, and mishandle pathology-specific terminology. Such gaps reinforce the need for refinement and deployment strategies that clinicians can actually trust.

 All code, models are publicly available at \url{https://github.com/hrlblab/MedFoundationHub}

\section{NEW OR BREAKTHROUGH WORK TO BE PRESENTED}
In summary, our contributions are threefold:
\begin{itemize}
    \item \textbf{MedFoundationHub Toolkit.} A GUI-based, plug-and-play framework for secure, on-premise deployment of medical VLMs, supporting both Hugging Face and local registries with full Docker integration.
    \item \textbf{Pathologist-in-the-Loop Evaluation.} First systematic assessment of multiple frontier VLMs by practicing board-certified pathologists within a controlled, privacy-preserving deployment environment.
    \item \textbf{Clinically Oriented Benchmark Analysis.} Comprehensive results on two pathology QA datasets (colon and renal), demonstrating both the potential and the current limitations of medical VLMs in real diagnostic scenarios.
\end{itemize}

% 背景+ Related Work(vlm/llm/deep learning application +Contrast Learning从文本提取embedding路线(Clip/Plip/Conch)+ Pathology/radiology captioning model(Prism,, Medgemma3）+LLM&docker部署+

\begin{figure}[htbp] \centering \includegraphics[width=0.93\linewidth]{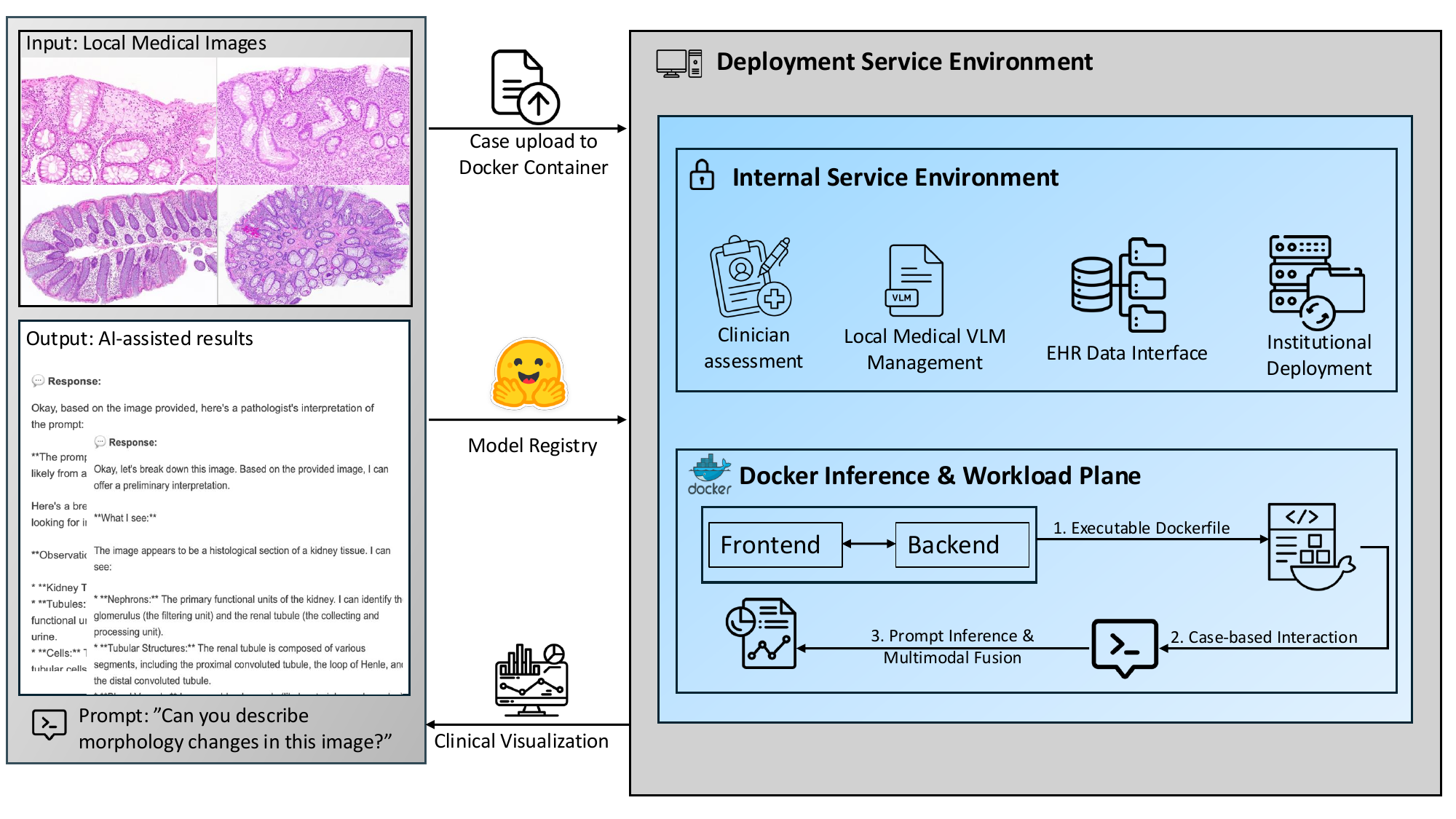} 
\caption{System architecture illustrating the isolation of the secret environment for sensitive data and inference, and the external environment for secure model management.}
\label{fig:system_architecture} 
\end{figure} 

\section{Methods}
MedFoundationHub is designed as a secure and extensible platform that enables institutional deployment and systematic evaluation of medical vision–language models (VLMs). The system follows a frontend–backend decoupled architecture in which clinicians interact through a web dashboard, while all model acquisition, inference, and logging occur in containerized backend services. Clinical inputs are formulated as natural-language prompts $q \in \mathcal{Q}$ accompanied by whole-slide pathology images $I \in \mathbb{R}^{H \times W \times C}$. 
For example:
\begin{quote}
\textit{“Can you describe morphology changes in this image?”}
\end{quote}
Given such a pair $(I, q)$, the system executes multimodal inference entirely within an on-premise environment. Each model $M_\theta$ runs inside an isolated Docker container, producing a structured output $y \in \mathcal{Y}$ according to
\[
y = M_\theta \big( f_{\text{img}}(I), f_{\text{text}}(q) \big),
\]
where $f_{\text{img}}$ and $f_{\text{text}}$ denote image and text encoders, respectively. By confining installation, execution, and storage to the institutional infrastructure, MedFoundationHub enforces strict data sovereignty and minimizes attack surface.
\vspace{0.3em}

\noindent \textbf{Model Management.}  Model management is supported through a dual registry design that accommodates both open-source models retrieved from Hugging Face and institution-specific local models. Retrieval occurs through a controlled external channel that downloads model weights $\theta$ only, after which all inference remains confined to the internal environment. Registered models are automatically containerized, ensuring reproducible deployment and version control. Formally, the registry maintains a set
\[
\mathcal{R} = \{ (M_{\theta_i}, v_i, \text{meta}_i) \}_{i=1}^N,
\]
where $v_i$ denotes the version tag and $\text{meta}_i$ includes provenance information and access logs. This registry enables seamless switching between VLMs, supporting comparative evaluation while guaranteeing that every inference event is logged for auditability. Extensible interfaces support integration with electronic health record (EHR) systems.

\vspace{0.3em}
\noindent \textbf{Inference and Workload Plane.}  
Inference is executed within the Docker inference and workload plane, which abstracts heterogeneous VLMs into a unified service interface. Each container implements a standardized API for multimodal fusion, where cross-attention modules combine encoded features:
\[
z = \text{CrossAttn}\big( f_{\text{img}}(I), f_{\text{text}}(q) \big), \quad
y = g(z),
\]
with $g(\cdot)$ mapping to task-specific outputs such as captions or multiple-choice answers. Hot deployment mechanisms allow models to be pulled, updated, and restarted dynamically without disrupting ongoing services, while autoscaling and batching stabilize latency under varying clinician workloads. Container-level telemetry—including GPU utilization $u_{gpu}(t)$, memory footprint $m(t)$, and latency distribution $L(\tau)$—is continuously collected, providing reproducible performance metrics for each deployment condition.

\vspace{0.3em}
\noindent \textbf{Clinician Interaction.}  
Clinician interaction occurs through a case-centric dashboard that integrates model outputs with visual evidence and supports expert-driven evaluation. Physicians can toggle between available models $M_{\theta_i}$, compare predictions $\{y_i\}$ side by side, and provide structured feedback $s \in \{0,1,2,3,4\}$ (see Table~\ref{tab:scoring}) . These annotations are stored in parallel with system logs. This interactive loop transforms raw VLM outputs into clinically interpretable insights while simultaneously generating high-value ground truth for benchmarking. By coupling containerized deployment with physician-in-the-loop evaluation, MedFoundationHub establishes an end-to-end workflow that enables secure institutional deployment of medical VLMs, systematic performance assessment, and extensible integration into real-world diagnostic practice.

\begin{figure}[htbp]
  \centering
  % 第一张图
  \begin{subfigure}[t]{0.98\linewidth}
    \centering
    \includegraphics[width=\linewidth]{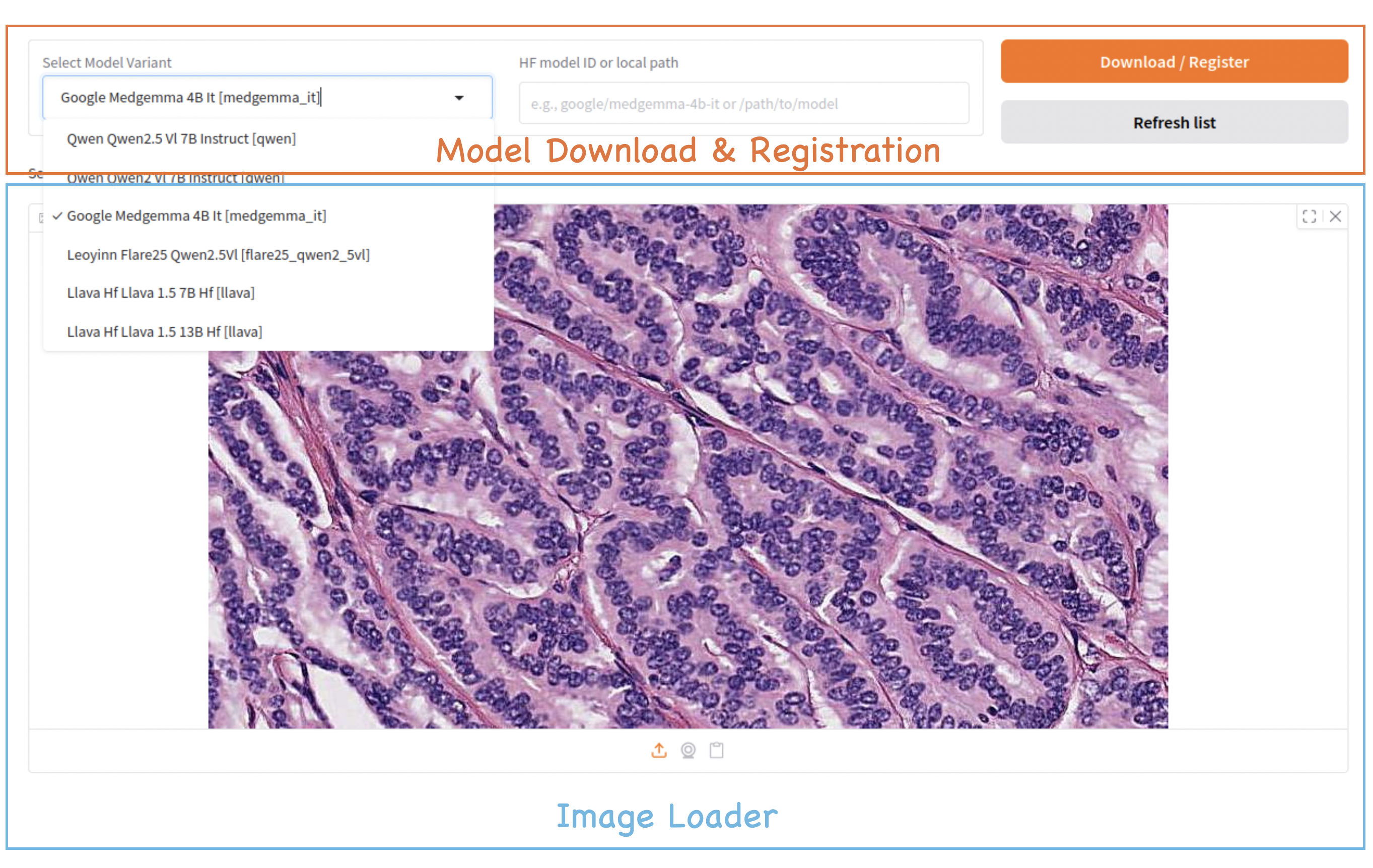}
    \caption{Model download and registration module. 
    Clinicians or engineers can add new VLMs by entering a Hugging Face model ID or a local path, after which the model is securely containerized and registered in the system.}
    \label{fig:ui1}
  \end{subfigure}
  \vspace{0.5 cm} % 调整上下间距

  % 第二张图
  \begin{subfigure}[t]{0.98\linewidth}
    \centering
    \includegraphics[width=\linewidth]{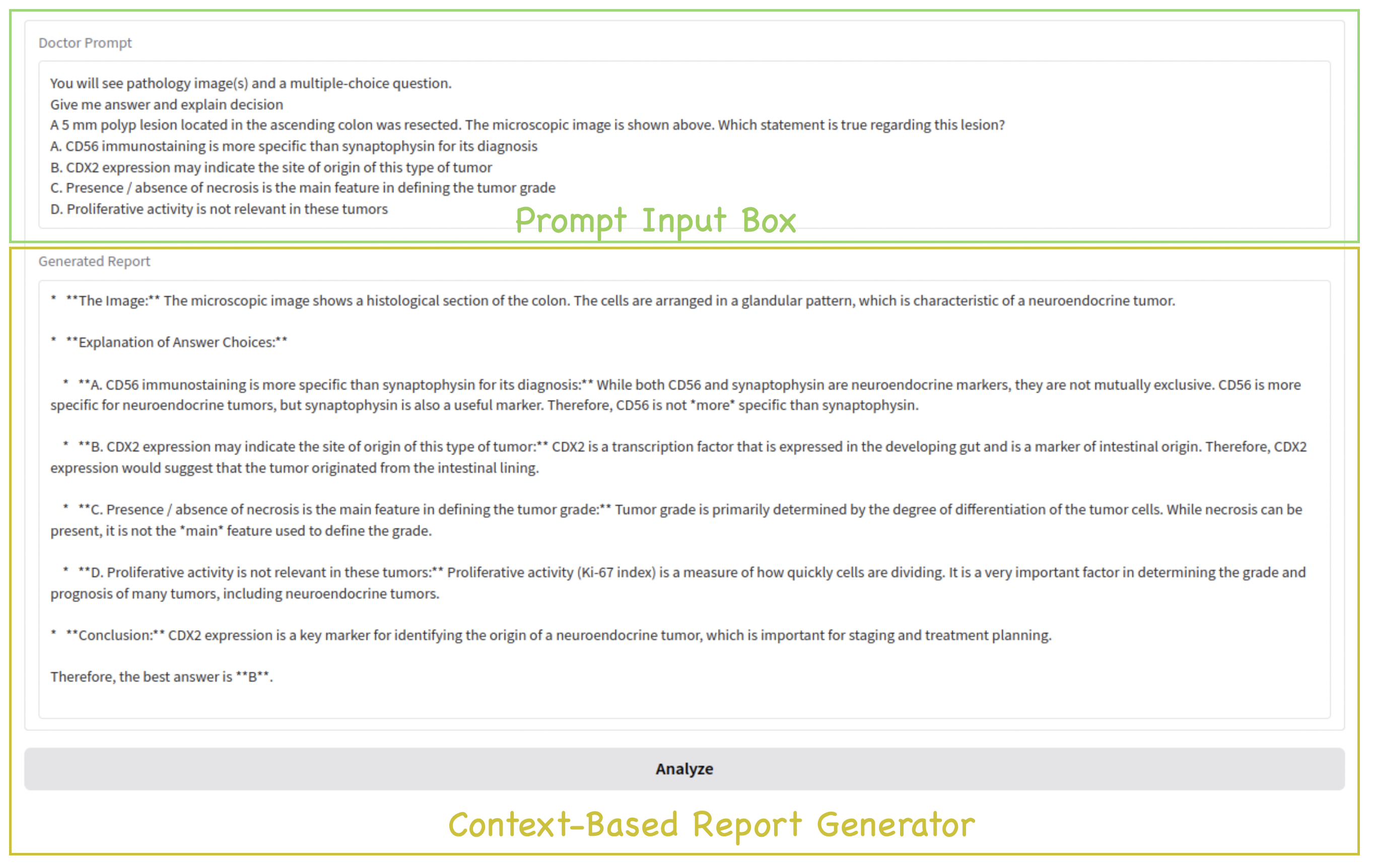}
    \caption{Prompt input and report generation module. 
    Users upload pathology images, enter free-text prompts, and invoke \textit{Analyze} to obtain reasoning-aware diagnostic outputs with context-based visualization.}
    \label{fig:ui2}
  \end{subfigure}

  \caption{System interface, illustrating (a) model acquisition/registration and (b) clinician-facing diagnostic interaction.}
  \label{fig:ui}
\end{figure}

Within the MedFoundationHub interface, clinicians begin by interacting with the \textbf{Model Download \& Registration} component, which allows selection of an available VLM $M_{\theta_i}$ from the registry or the addition of new models via Hugging Face identifiers or local file paths. When a new model is registered, the system automatically retrieves its weights $\theta$, builds a container, and records the entry in the registry
\[
\mathcal{R} = \{ (M_{\theta_i}, v_i, \text{meta}_i) \}_{i=1}^N ,
\]
ensuring reproducibility and version control across deployments.  

Case data are introduced through the \textbf{Image Loader}, which accepts whole-slide pathology images $I \in \mathbb{R}^{H \times W \times C}$, and the \textbf{Prompt Text Box}, where clinicians articulate diagnostic questions or case descriptions $q \in \mathcal{Q}$. The paired input $(I,q)$ is encapsulated into a standardized case representation and routed to the backend. By pressing the \textit{Analyze} button, the toolkit automatically dispatches the request to the Dockerized inference service. Each container performs multimodal fusion of the encoded image and text features:
\[
z = \text{CrossAttn}\big( f_{\text{img}}(I), f_{\text{text}}(q) \big), 
\quad y = g(z),
\]
with $y \in \mathcal{Y}$ denoting the generated output.  

Finally, results are presented in the \textbf{Context-Based Report Generator}, which formats raw model outputs into clinically interpretable artifacts—ranging from morphological descriptions to structured multiple-choice answers or explanatory reasoning chains. This module not only improves readability but also provides hooks for downstream auditing, structured logging, and side-by-side model comparison.  

Overall, the workflow reduces physician interaction to three intuitive steps: selecting or registering a model, uploading case data, and initiating inference. All remaining processes, including container orchestration, privacy enforcement, telemetry collection, and report formatting, are handled transparently by the toolkit. By making explicit each UI component (download \& registration, image loader, prompt input, and report generator), MedFoundationHub bridges the gap between backend technical infrastructure and the practical needs of physicians, offering a reproducible and institution-ready pipeline that minimizes cognitive load while preserving rigorous security guarantees.

\section{Experimental Setting}

The experimental design emphasizes \textbf{clinician-centered evaluation of the MedFoundationHub system}, rather than large-scale benchmarking of individual models. Board-certified pathologists interacted directly with the deployed platform, using it as they would in a practical diagnostic support workflow. Five representative VLMs were made available within the system registry, enabling clinicians to toggle between them through the interface. Table~\ref{tab:model_overview} summarizes their approximate parameter sizes and the rationale for inclusion. These models span a spectrum of capacities from the compact MedGemma3-4B, which balances medical reasoning and context length with asset-limited deployment, to the larger LLaVA-1.5-13B for evaluating scale effects in complex visual reasoning. Including both the Qwen2 and Qwen2.5 Instruct variants allows us to capture the impact of architectural optimization and instruction tuning that recent models offer.

\begin{table}[htbp]
\centering
\caption{Overview of the VLMs evaluated within MedFoundationHub: parameter sizes and rationale for inclusion.}
\label{tab:model_overview}
\begin{tabular}{lp{2.5cm}p{7cm}}
\toprule
\textbf{Model} & \textbf{Approx. Size} & \textbf{Rationale for Selection} \\
\midrule
Google-MedGemma3-4B  & $\sim$4B & Compact multimodal model with 128K-token context and strong medical reasoning capabilities \\
Qwen2-VL-7B-Instruct     & $\sim$7B & Balanced size with instruction tuning, structured output improvements \\
Qwen2.5-VL-7B-Instruct   & $\sim$7B & Latest Qwen2.5 improvements: long-text generation, structured output efficiency, suitable resource footprint \\
LLaVA-1.5-7B          & $\sim$7B & Fine-tuned for visual instruction tasks, early multimodal baselines with strong VQA performance \\
LLaVA-1.5-13B         & $\sim$13B & Larger-capacity variant to assess scaling effects in multimodal reasoning; supports higher visual fidelity \\
\bottomrule
\end{tabular}
\end{table}

The colon pathology dataset was obtained from PathologyOutlines, while the renal pathology dataset was retrieved from Arkana Labs, the AJKD Atlas of Renal Pathology, NephSim and other sources. Within MedFoundationHub, board-certified pathologists engaged with every case from both sources across all five models, resulting in \textit{1,015 independent scoring events per pathologist} (Table~\ref{tab:datasets}). These figures reflect the volume of structured expert annotations generated through the platform rather than an attempt to exhaustively benchmark model accuracy. By framing the study as \textbf{clinician-in-the-loop evaluation}, we emphasize the workload and consistency of human scoring, which constitutes the core evidence for our analysis.  

\begin{table}[htbp]
\centering
\caption{Pathologist evaluation workload within MedFoundationHub. Each case was reviewed across all models, producing a total of 1{,}015 structured expert scores.}
\label{tab:datasets}
\begin{tabular}{lccc}
\toprule
\textbf{Dataset} & \textbf{Domain} & \textbf{Cases per Dataset} & \textbf{Expert Scores (5 Models)} \\
\midrule
Colon pathology dataset & Colon pathology & 98  & 490 \\
Renal pathology dataset        & Renal pathology & 105 & 525 \\
\midrule
\textbf{Total}             & ---             & 203 & 1{,}015 \\
\bottomrule
\end{tabular}
\end{table}

Clinician scoring followed a five-level rubric (0–4) designed to capture both correctness and reasoning fidelity. The dashboard enabled experts to compare outputs across models, annotate reasoning quality, and log case-level evaluations. This interactive loop generated a corpus of structured clinician feedback coupled with container-level telemetry (e.g., latency, memory, GPU utilization), thereby providing an integrated view of both \textbf{system usability} and \textbf{model behavior under deployment}. By design, the study measures how physicians would experience and assess the toolkit in practice, emphasizing usability, interpretability, and auditability, rather than dataset-driven performance benchmarks.

\begin{table}[t] 
\centering 
\caption{Clinician scoring rubric for model outputs.} 
\label{tab:scoring} 
\begin{tabular}{clp{7.5cm}} 
\toprule \textbf{Score} & \textbf{Definition} & \textbf{Rationale} 
\\ \midrule 0 & No answer & The model failed to produce any interpretable output; reflects inability to engage with the query. 
\\ 1 & Wrong answer & Output is clearly incorrect or contradicts established pathology knowledge; ensures penalization of unsafe hallucinations. 
\\ 2 & Partially correct answer & Output captures some relevant features but omits or misinterprets key diagnostic elements; highlights incomplete clinical utility. 
\\ 3 & Correct answer with wrong reasoning & Final prediction is correct, but explanation or morphological reasoning is flawed; stresses importance of interpretability and reasoning consistency. 
\\ 4 & Correct answer with correct reasoning & Both the diagnostic conclusion and reasoning process are clinically sound; represents the target level of safe, trustworthy performance. 
\\ \bottomrule \end{tabular} 
\end{table}

\section{Results}

\begin{figure}[htbp]
  \centering

  % 上面的直方图
  \begin{subfigure}[t]{0.75\linewidth}
    \centering
    \includegraphics[width=\linewidth]{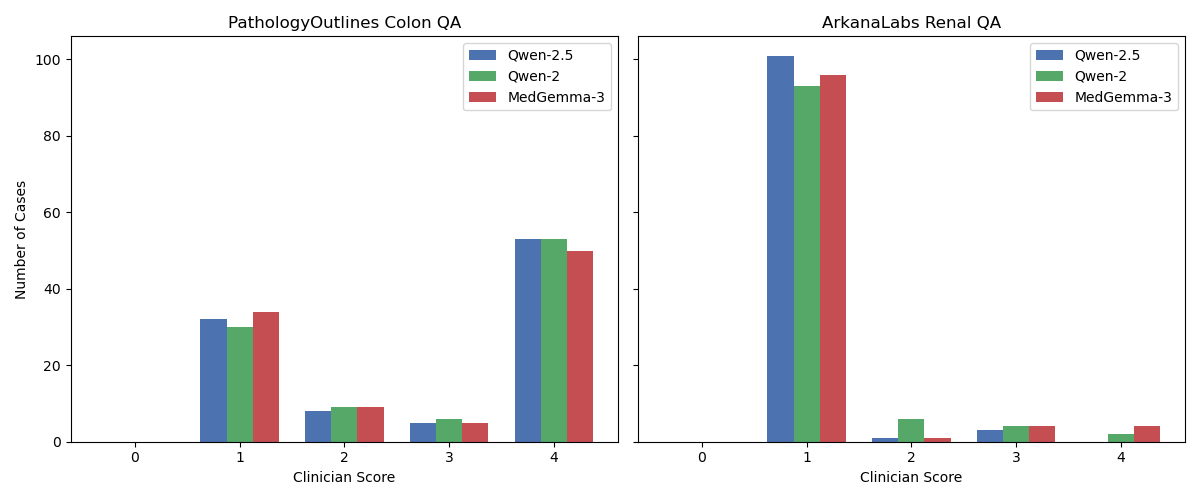}
    \caption{Distribution of clinician-assigned scores across models.}
    \label{fig:colon_case_hist}
  \end{subfigure}

  \vspace{0.4cm} % 上下间隔，可调整

  % 下面的百分比堆叠图
  \begin{subfigure}[t]{0.75\linewidth}
    \centering
    \includegraphics[width=\linewidth]{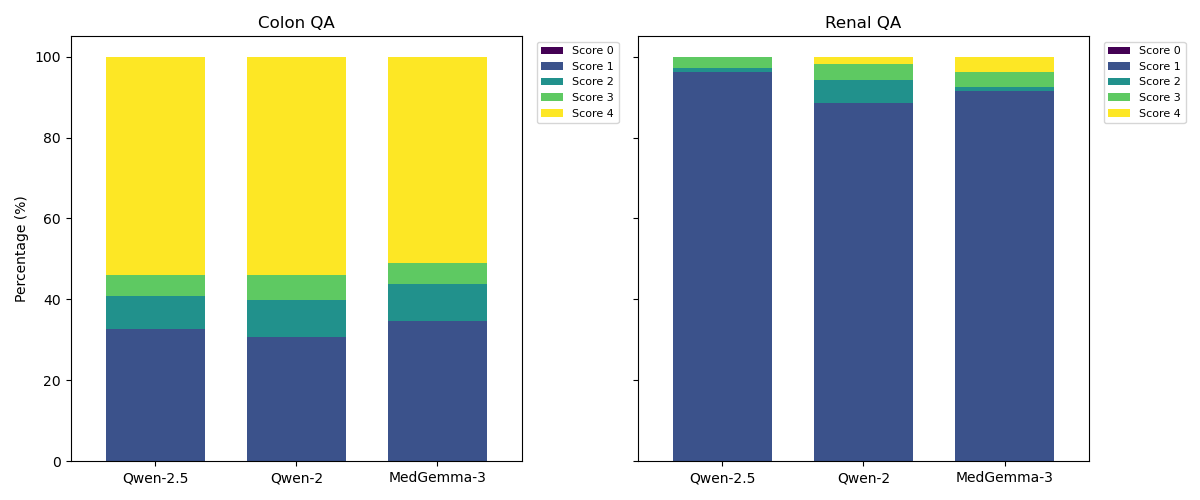}
    \caption{Stacked percentage representation of score distributions.}
    \label{fig:colon_case_percentage}
  \end{subfigure}

  \caption{Clinician evaluation of model outputs, shown as (a) absolute distribution and (b) normalized percentage distribution across scoring categories.}
  \label{fig:colon_case_examples}
\end{figure}

\subsection{Qualititive Results}
Although five VLMs were deployed within MedFoundationHub, only three: \textit{Qwen2-7B}, \textit{Qwen2.5-7B}, and \textit{MedGemma-3-4B}—were subjected to detailed clinician scoring. The rationale is that \textit{LLaVA-1.5-7B} and \textit{LLaVA-1.5-13B}, while valuable as multimodal baselines, predominantly produce short, direct answers without explicit reasoning chains. Since our rubric (\S\ref{fig:colon_case_examples}) requires assessment of both diagnostic correctness and reasoning fidelity, these models could not be consistently mapped to the 0–4 scale, and were therefore excluded from quantitative scoring to maintain fairness and methodological clarity.
For the dataset (98 cases per model), Qwen2-7B and Qwen2.5-7B achieved broadly similar distributions, with roughly half of cases receiving top scores (53/98 and 53/98, respectively), while MedGemma-3-4B trailed slightly (50/98). Misclassifications were concentrated in the 1–2 range, reflecting partial recognition of morphologic features but inaccurate attribution. In contrast, performance deteriorated sharply on the dataset (105 cases per model). Here, nearly all outputs from Qwen2.5-7B were scored as 1 (101/105), with only four instances rising above partial correctness. Qwen2-7B and MedGemma-3-4B exhibited modestly better balance, with a handful of correct answers and reasoning chains (scores 3–4), but the overwhelming majority of cases still clustered in the “wrong” category.

Clinician feedback corroborated these quantitative findings. By contrast, both LLaVA models produced terse answers with minimal reasoning, rendering them unsuitable for rubric-based scoring. Experts also noted peculiar behaviors: Qwen2.5-7B frequently returned a lettered choice (e.g., “C” or “F”) before providing any descriptive explanation, a format incongruent with standard diagnostic reporting. Additionally, LLaVA-1.5-7B occasionally generated truncated answers that appeared cut off mid-sentence, raising concerns about model stability in practical deployment.

\begin{figure}[htbp]
\begin{subfigure}[t]{0.49\linewidth}
    \centering
    \includegraphics[width=\linewidth]{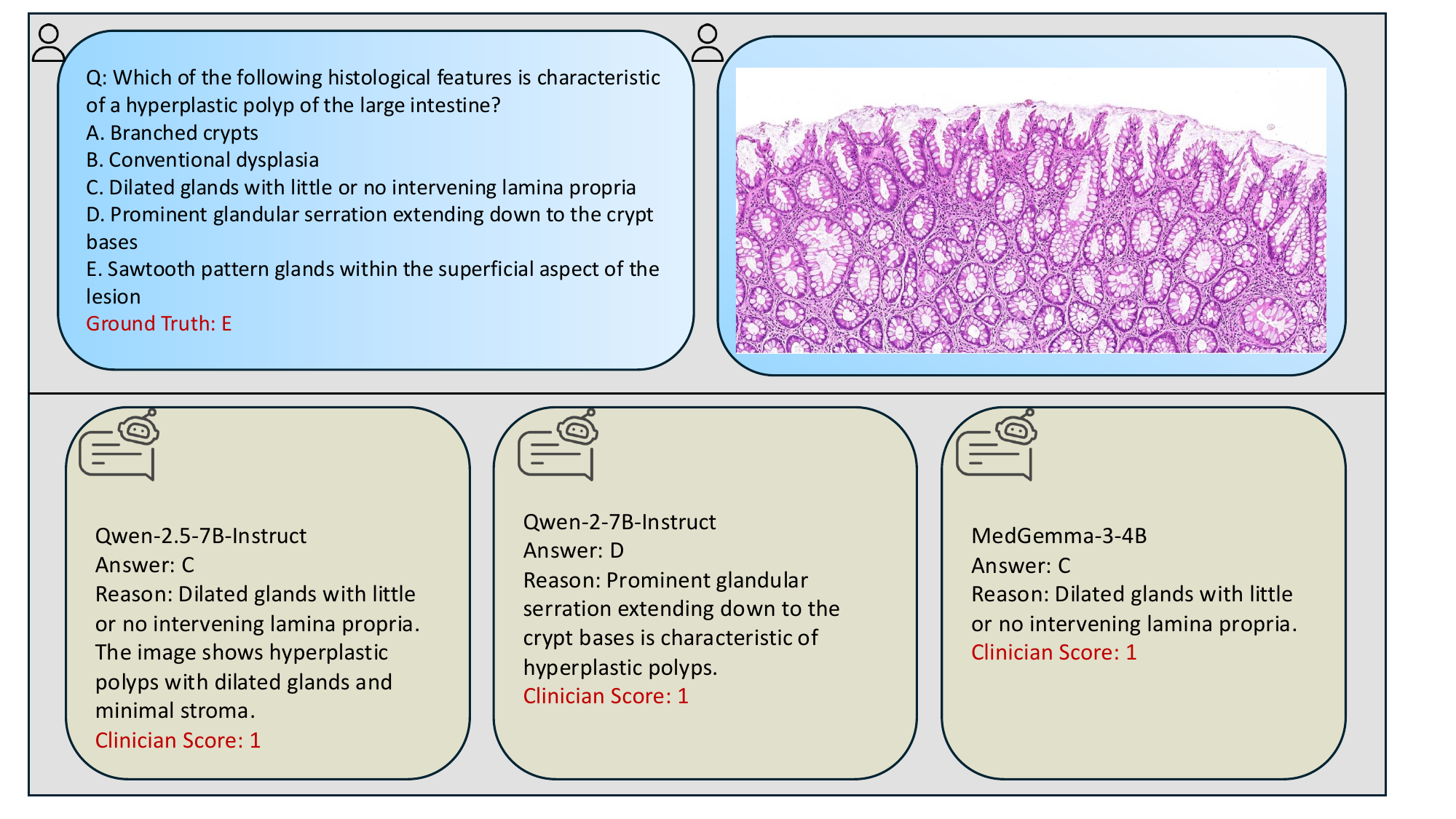}
    \caption{Case A: Colon QA example (score = 1).}
    \label{fig:colon_case_correct}
  \end{subfigure}
    \hfill
    \begin{subfigure}[t]{0.49\linewidth}
    \centering
    \includegraphics[width=\linewidth]{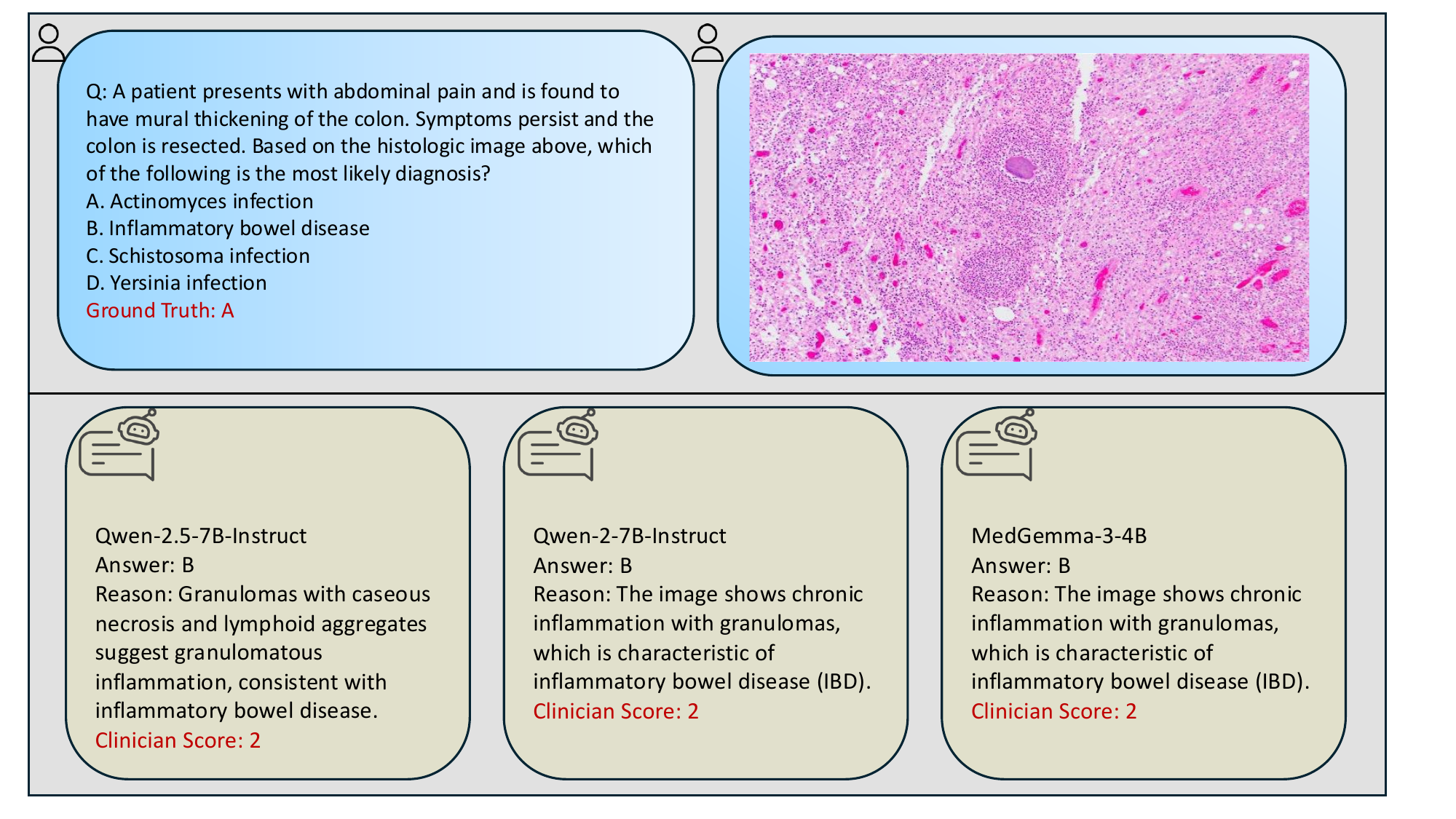}
    \caption{Case B: Colon QA example (score = 2).}
    \label{fig:colon_case_correct}
  \end{subfigure}
    \hfill
  \begin{subfigure}[t]{0.49\linewidth}
    \centering
    \includegraphics[width=\linewidth]{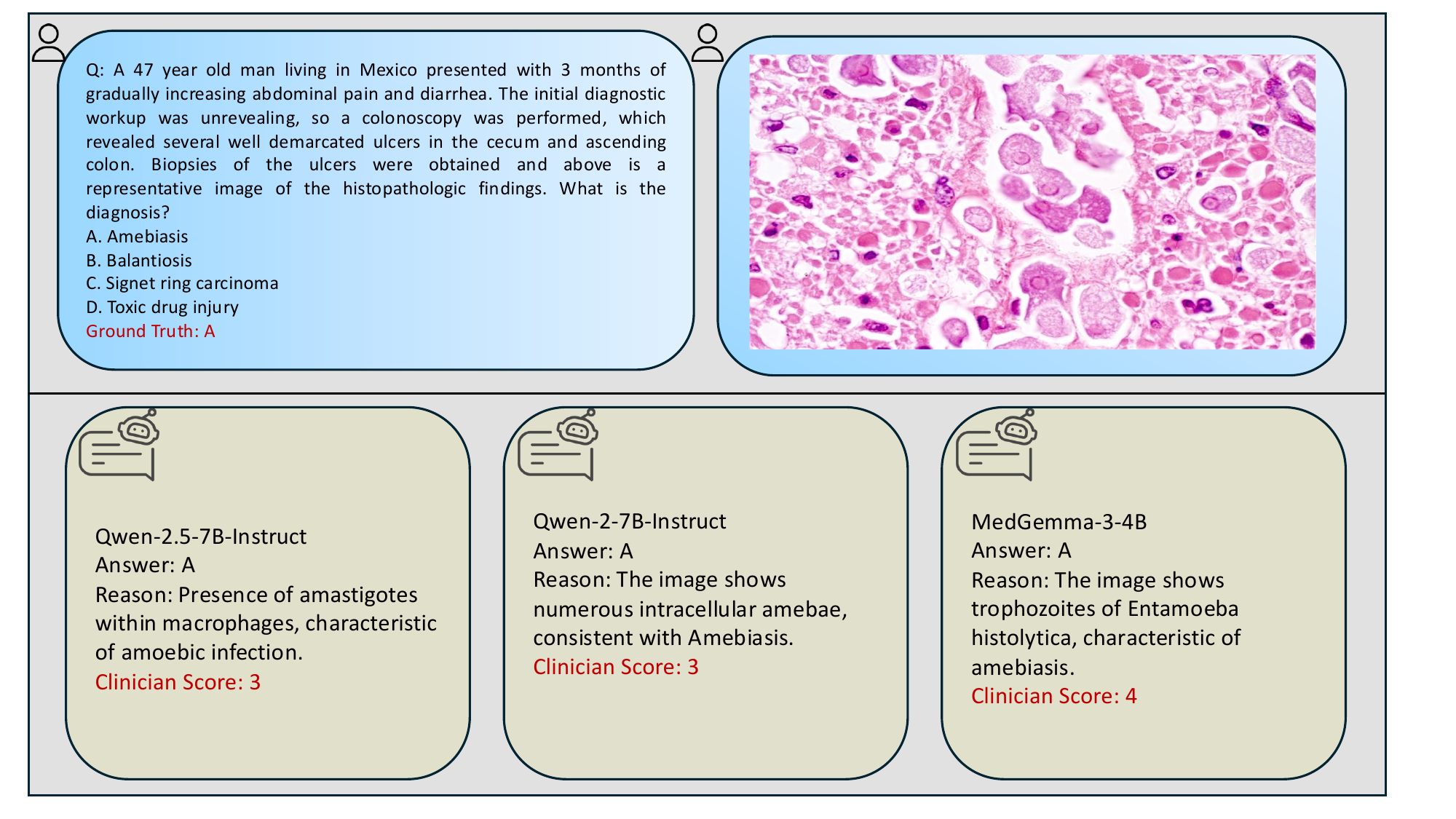}
    \caption{Case C: Colon QA example (score = 3 vs 4).}
    \label{fig:colon_case_correct}
  \end{subfigure}
 \hfill
   \centering
  \begin{subfigure}[t]{0.49\linewidth}
    \centering
    \includegraphics[width=\linewidth]{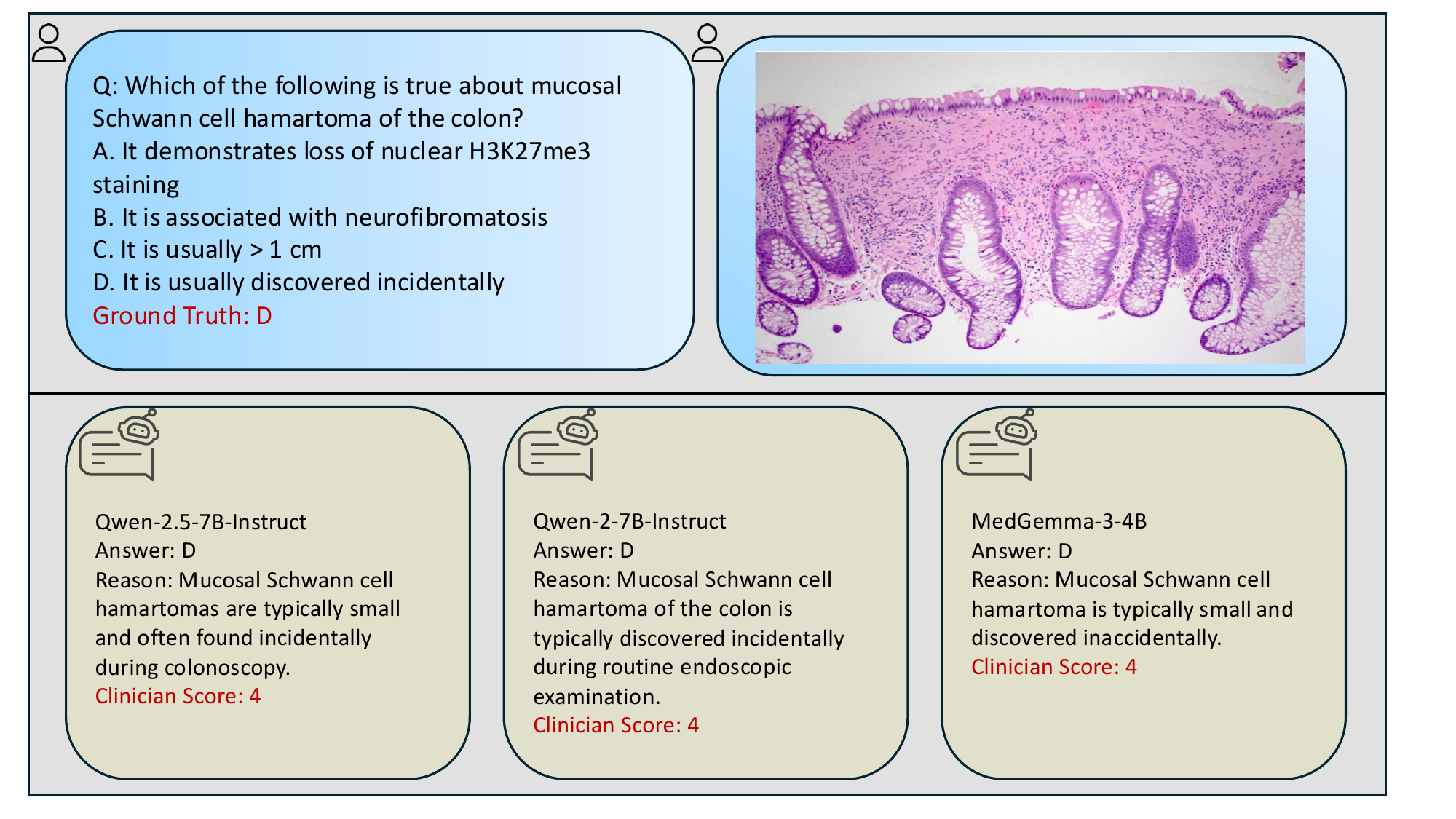}
    \caption{Case D: Colon QA example (score = 4).}
    \label{fig:colon_case_wrong}
  \end{subfigure}
  \caption{Representative case-based evaluations from \textit{PathologyOutlines Colon QA}.}
  \label{fig:colon_case_examples}
\end{figure}

\subsection{Case Studies}
We highlight four representative cases where model performance diverged from clinical expectations.

Case A (score = 1).
The task concerned the diagnosis of a hyperplastic polyp. The defining feature is a sawtooth serration limited to the superficial crypt epithelium, without basal distortion or cytologic atypia. None of the models identified this hallmark. Instead, answers referred to changes such as basal crypt serration or dysplasia—features more consistent with sessile serrated lesions or adenomas. Although the language was confident, the morphologic anchor was incorrect, making the outputs misleading for practice. Pathologists scored this as 1, emphasizing that a model which misses the distribution of serration cannot reliably separate hyperplastic polyps from their malignant mimics.

Case B (score = 2).
A resection specimen with colonic mural thickening was queried. The correct diagnosis was Actinomyces infection, characterized by sulfur granules—dense basophilic bacterial colonies with eosinophilic club-like peripheries—within suppurative inflammation. All models instead reported inflammatory bowel disease, citing granulomas and chronic lymphoid aggregates. While chronic inflammation was indeed present, the models entirely missed the filamentous bacterial colonies, a pathognomonic finding. This partial recognition of background inflammation, but not the specific etiology, led experts to assign a score of 2.

Case C (score = 3–4).
A case of amebic colitis was tested. The ground truth was Amebiasis, recognizable by Entamoeba histolytica trophozoites with ingested red blood cells at the ulcer edge. All models gave the correct final answer. However, two responses only described “intracellular amebae” without discriminating from macrophages, which is a common pitfall in routine sign-out, and were thus scored 3. By contrast, one model explicitly identified trophozoites of Entamoeba histolytica and mentioned erythrophagocytosis, earning a score of 4 for both correctness and morphologic specificity.

Case D (score = 4).
The final case involved mucosal Schwann cell hamartoma. Hallmark features are small, intramucosal nodules of bland spindle cells, diffusely S100-positive, unassociated with syndromic disease, and most often discovered incidentally during colonoscopy. All evaluated models correctly described these points, highlighting incidental discovery and excluding neurofibromatosis. Pathologists therefore assigned a score of 4, affirming that both the conclusion and the reasoning aligned with established diagnostic standards.

\section{Conclusion and Discussion}
MedFoundationHub demonstrates that secure, modularized deployment of medical vision–language models is both technically feasible and clinically accessible. By decoupling frontend interaction from backend orchestration and enforcing strict containerized isolation, the system lowers adoption barriers for non-technical users while ensuring compliance with institutional privacy requirements. Beyond mere engineering convenience, the platform creates a reproducible environment in which clinicians can interrogate, compare, and critique frontier VLMs, generating domain-informed assessments that are rarely captured by conventional machine learning benchmarks.

Our experiments highlight that, despite promising progress, mainstream VLMs, whether general-purpose or medically adapted, still fall short in real diagnostic contexts. Limitations include off-target or vague responses, superficial reasoning chains, and insufficient mastery of domain-specific terminology. These shortcomings underscore that medical deployment cannot rely solely on scaling general models, but instead demands domain-customized training, multimodal alignment strategies, and integration with curated electronic health record (EHR) data.

Looking forward, we envision MedFoundationHub not only as a deployment toolkit but as an evolving testbed for bridging AI and pathology practice. Future directions include: expanding compatibility with larger and more diverse medical VLMs, incorporating advanced visualization modules to better capture morphology–text interactions, and conducting longitudinal user studies with practicing pathologists. By embedding clinical expertise directly into the evaluation loop, we aim to accelerate the development of safe, interpretable, and trustworthy AI systems that can be integrated into everyday diagnostic workflows. Ultimately, MedFoundationHub aspires to transform from a research prototype into an institutional infrastructure for the responsible adoption of multimodal AI in healthcare.

\acknowledgements
% equivalent to \section*{ACKNOWLEDGMENTS}       
\begin{flushleft}
This research was supported by NIH R01DK135597 (Huo), DoD HT9425-23-1-0003 (HCY), and KPMP Glue Grant. This work was also supported by Vanderbilt Seed Success Grant, Vanderbilt Discovery Grant, and VISE Seed Grant. This project was supported by The Leona M. and Harry B. Helmsley Charitable Trust grant G-1903-03793 and G-2103-05128. This research was also supported by NIH grants R01EB033385, R01DK132338, REB017230, R01MH125931, and NSF 2040462. We extend gratitude to NVIDIA for their support by means of the NVIDIA hardware grant. This work was also supported by NSF NAIRR Pilot Award NAIRR240055.
\end{flushleft}
\bibliographystyle{spiebib} 
\bibliography{report.bib}

\end{document}